\documentclass{bmcart}
\pdfoutput=1

\usepackage[utf8]{inputenc} 

\usepackage{enumerate}
\usepackage{url}
\usepackage{multirow}
\usepackage{rotating}
\usepackage{makecell}
\usepackage{listings}
\usepackage[acronym,nonumberlist]{glossaries}

\makeglossaries

\newacronym{kb}{KB}{knowledge base}
\newacronym{lod}{LOD}{Linked Open Data}
\newacronym{yago}{YAGO}{Yet another great ontology}
\newacronym{umls}{UMLS}{Unified Medical Language System}
\newacronym{rdf}{RDF}{Resource Description Framework}
\newacronym{ctgov}{CTGov}{ClinicalTrials.gov}
\newacronym{eudra}{EudraCT}{European Clinical Trials Database}
\newacronym{ictrp}{ICTRP}{WHO International Clinical Trials Registry Platform}
\newacronym{iri}{IRI}{Internationalized Resource Identifier}
\newacronym{mesh}{MeSH}{Medical Subject Headings}
\newacronym{sparql}{SPARQL}{SPARQL Protocol and RDF Query Language}
\newacronym{csv}{CSV}{Comma-separated values}
\newacronym{tfidf}{TF-IDF}{term frequency–inverse document frequency}
\newacronym{rdfs}{RDFS}{Resource Description Framework Schema}
\newacronym{skos}{SKOS}{Simple Knowledge Organization System}
\newacronym{who}{WHO}{World Health Organization}

\startlocaldefs
\newcommand{\specialcell}[2][c]{%
  \begin{tabular}[#1]{@{}l@{}}#2\end{tabular}}

\lstdefinelanguage{sparql}{
    morecomment=[l]{\#},
    language=SQL,
    morekeywords={SELECT, PREFIX},      
    commentstyle=\itshape\color{blue},
    sensitive=true,
}
\endlocaldefs

\begin{document}

\begin{frontmatter}

\begin{fmbox}
\dochead{Research}


\title{Access to care: analysis of the geographical distribution of healthcare using Linked Open Data}


\author[
   addressref={aff1, aff2},               
   corref={aff1},                   
   noteref={n1},                    
   email={s.baezsantamaria@vu.nl}   
]{\inits{S}\fnm{Selene} \snm{B\'{a}ez Santamar\'{i}a}}
\author[
   addressref={aff2},
   noteref={n1},
   email={manousogm@gmail.com}
]{\inits{E}\fnm{Emmanouil} \snm{Manousogiannis}}
\author[
   addressref={aff1,aff2},
   noteref={n1},
   email={guusjeboomgaard@hotmail.com}
]{\inits{G}\fnm{Guusje} \snm{Boomgaard}}
\author[
   addressref={aff1,aff2},
   noteref={n1},
   email={trplinh@gmail.com}
]{\inits{LP}\fnm{Linh P.} \snm{Tran}}
\author[
   addressref={aff2},
   email={zoltan.szlavik@mytomorrows.com}
]{\inits{Z}\fnm{Zolt\'{a}n} \snm{Szl\'{a}vik}}
\author[
   addressref={aff2},
   email={r.sips@mytomorrows.com}
]{\inits{RJ}\fnm{Robert-Jan} \snm{Sips}}


\address[id=aff1]{
  \orgname{Vrije Universiteit Amsterdam}, 
  \street{De Boelelaan 1111},                     %
  \postcode{1081 HN}            
  \city{Amsterdam},                  
  \cny{The Netherlands}              
}
\address[id=aff2]{%
  \orgname{MyTomorrows},
  \street{Anthony Fokkerweg 61},
  \postcode{1059 CP}
  \city{Amsterdam},
  \cny{The Netherlands}
}


\begin{artnotes}
\note[id=n1]{Work produced while affiliated at MyTomorrows} 
\end{artnotes}

\end{fmbox}


\begin{abstractbox}

\begin{abstract} 
\textbf{Background}: Access to medical care is strongly dependent on resource allocation, such as the geographical distribution of medical facilities. Nevertheless, this data is usually restricted to country official documentation, not available to the public. While some medical facilities' data is accessible as semantic resources on the Web, it is not consistent in its modeling and has yet to be integrated into a complete, open, and specialized repository. This work focuses on generating a comprehensive semantic dataset of medical facilities worldwide containing extensive information about such facilities' geo-location. 
\newline
\textbf{Results}: For this purpose, we collect, align, and link various open-source databases where medical facilities' information may be present. This work allows us to evaluate each data source along various dimensions, such as completeness, correctness, and interlinking with other sources, all critical aspects of current knowledge representation technologies.
\newline
\textbf{Conclusions}: Our contributions directly benefit stakeholders in the biomedical and health domain (patients, healthcare professionals, companies, regulatory authorities, and researchers), who will now have a better overview of the access to and distribution of medical facilities. 

\end{abstract}




\end{abstractbox}
%

\end{frontmatter}



\section*{Background} 
\label{sec:intro}
Equal access to healthcare is a fundamental, irrevocable human right and is protected by international law \cite{UNComment14}. Access to primary healthcare has been studied along various dimensions, such as the geographic spread of specialized healthcare sites \cite{anderson2014breast,lin2016association}, racial and/or ethnic factors \cite{manuel2018racial,peipins2013racial} and gender \cite{pinillos2018examining}. The aforelisted confirm that despite legislation and good intentions, healthcare is not universally accessible. 

Yet another factor in determining the quality of healthcare is access to novel treatments. Several studies indicate that participation in clinical trials can result in improved health conditions and lower-mortality rates in rare or life-threatening conditions \cite{research_participation,women_in_trials,breast_cancer_trials}. However, the above is not the only reason why analyzing the geo-location of clinical research is of significant importance. Genome-related diseases do not spread equally across ethnicity and geography \cite{tan2016cancer}; while diseases that were bound to a specific geographical location are nowadays spreading more widely due to climate change and globalization \cite{world2013sustaining}.  In some cases, effectively targeting the right population might provide a solution to the issues hampering healthcare research. For example, Stensland \textit{et al.} \cite{stensland2014adult} estimate that about 39\% of cancer trials fail because of low accrual, and only $\approx 5\%$ of patients referred are eventually able to access the trial.
%

Despite the importance of monitoring and studying (disparities in) access to healthcare, we lack a comprehensive \acrfull{kb} containing medical facilities that offers the following: a) high geographic coverage (e.g., cover various countries), b) complete location information (e.g., geo-coordinates), c) validity (e.g., excludes outdated or inaccurate information), and d) rich interlinking with external data sources for machine-readability. The World Health Organization reports related aggregated metrics, for example, the number of hospitals per 100,000 people in a given country \cite{whoreport}; yet their latest report refers to information from 2014. Meanwhile, other open-source initiatives \cite{healthsites} have focused on curating and visualizing data from a single source (i.e. OpenStreetMaps), which inherently contains biased data. Detailed medical facility location data is currently distributed and does not come in a standardized and machine-recognizable format, thus making it challenging to explore, use, and analyze. 

\vspace{0.2cm}
\noindent In this paper, i) we evaluate the coverage and quality of various data sources with regards to medical facilities and their geo-location, and provide recommendations for future use of these data sources, especially the \acrfull{lod} repositories; ii) we create an aligned database of worldwide medical facilities and illustrate its potential to investigate access to healthcare by analyzing the facilities' geographical distribution; ii) we perform simple entity linking to match clinical trial sites to medical facilities in the aforementioned database and we further investigate the hurdles with regards to geographical search of pre-approval treatments.

\subsection*{Related Work} 
\label{sec:related_work}
\subsection*{Linked Open Data} 
The idea of \acrshort{lod} came with the purpose of making data accessible and interconnected across sources and domains. The \acrshort{lod} Cloud collects all datasets that are present and interlinked on the Web. As of March 2020, the \acrshort{lod} Cloud contained 1,260 datasets and 16,187 links.\footnote{\url{https://lod-cloud.net/#about}}. 
 Farber \textit{et al.} made an in-depth study comparing five of the largest open and accessible knowledge graphs: DBpedia, Freebase, OpenCyc, Wikidata, and \acrfull{yago} \cite{farber2018linked}. The study defines dimensions along which we can evaluate a KB and metrics to assess it quantitatively. Among these, we may emphasize interlinking, defined as 'the extent to which entities representing the same concept are linked to one another'.

Previous work has explored the intersection between \acrshort{lod} and biomedical sciences\footnote{\url{https://lod-cloud.net/clouds/life-sciences-lod.svg}}. 
For example, Cuzzola \textit{et al.} \cite{cuzzola2018umls} worked on linking domain-specific \acrshort{kb}s like the \acrfull{umls} \cite{bodenreider2004unified} to general-domain \acrshort{lod} resources like DBpedia; while Zehra \textit{et al.} \cite{zehra2019cancer} focused their efforts on datasets in the sub-domain of cancer genomics. Some of these initiatives illustrate the impact \acrshort{lod} resources may have on improving access to health care. Hassanzadeh \textit{et al.} \cite{hassanzadeh2015automatic} created an \acrfull{rdf} representation of \acrfull{ctgov}\footnote{\url{https://clinicaltrials.gov/}}  to allow the inclusion of this resource to the \acrshort{lod} landscape. Bukhari \textit{et al.} \cite{bukhari2013canadian} published the Canadian health census as Linked Open Data. Zaveri \textit{et al.} \cite{zaveri2011redd} focused on quantifying disease-research disparity using open Web resources. 

Special attention to geographical information in the \acrshort{lod} Cloud has been explored in use cases such as delineation of functional urban areas \cite{khalili2018using} and identification of locations in historical data \cite{beek2018nlgis}.

\subsubsection*{Treatment search} 
Searching for treatments in development remains a challenging and time-consuming task since the current tools for searching clinical trials are sub-optimal \cite{korkontzelos2011text}. In the context of this work, we observe that these interfaces are not flexible or usable when it comes to searching the hospitals involved in clinical trials. For example, \acrshort{ctgov} allows users to search by country or alternatively by location keywords in their advanced search; the \acrfull{eudra}\footnote{\url{https://www.clinicaltrialsregister.eu/ctr-search/search}} advanced search only allows to filter by country; while the \acrfull{ictrp}\footnote{\url{http://apps.who.int/trialsearch/default.aspx}} does not support geographic search. 


\section*{Methods} 
\label{sec:medical-facilities}
\subsection*{Requirements for Knowledge Base Construction} 
To serve the use case of geography-centered search for (approved or pre-approved) treatment options, we set the goal to create a KB that gathers location information for medical facilities. The requirements for our \acrshort{kb} are as follows: 

\begin{enumerate}[\hspace{.5cm}\textbf{Requirement} 1\textbf{:}]
    \item To gather healthcare facilities and potential clinical trial sites.
    \item To gather the entities' names and alternative names in various languages.
    \item To gather the entities' location details (country, state, city, zip, latitude, and longitude coordinate pairs) as semantic resources themselves.
    \item To validate the entities' properties as valid locations and names.
\end{enumerate}

\noindent To generate such a \acrshort{kb} we follow the steps below:
\begin{enumerate}[\hspace{.5cm}\textbf{Step }1\textbf{:}]
    \item Relevant data source and resource identification 
    \item Data querying and gathering
    \item Resource and ontology alignment 
\end{enumerate}

\vspace{0.2cm}
To satisfy \textbf{Requirement 1}, we identify three different categories for potential trial sites: hospitals, universities, and pharmaceutical companies. These are sites where clinical research and trials may take place. To address these categories in more detail, the category of hospitals in our case includes clinics and research institutes that have any relation to diseases. Furthermore, the concept of a university is not the same in every country. Therefore, we broaden this category to higher educational institutions. Finally, pharmaceutical companies include businesses in the pharmaceutical industry, biotechnology, and healthcare industry. An important notion here is that there could be overlaps between the categories. 

To satisfy \textbf{Requirement 2}, we include labels and alternative labels of the locations in at least 30 different languages. During the exploration of the data sources, we noted that most of them link to semantic resources from the GeoNames vocabulary. Therefore, to satisfy \textbf{Requirement 3} we collect location data mapped to GeoNames when possible and attempt to link other string representations of locations to the GeoNames vocabulary.

Finally, to satisfy \textbf{Requirement 4} we implement a `blacklisting' strategy to avoid retrieving invalid resources such as broad or abstract concepts. 

\subsection*{Data sources} 
We integrate four different LOD resources: DBpedia \cite{auer2007dbpedia}\footnote{\url{http://dbpedia.org/}}, Wikidata \cite{vrandevcic2012wikidata}\footnote{\url{https://www.wikidata.org/}}, \acrshort{yago}2 \cite{hoffart2013yago2}\footnote{\url{https://www.mpi-inf.mpg.de/departments/databases-and-information-systems/research/yago-naga/yago/}}, and GeoNames \cite{ahlers2013assessment}\footnote{\url{https://www.geonames.org/}}. We also use the \acrshort{umls}, due to its significance in the biomedical domain. We disregard \acrshort{lod} resources that would pose a challenge in terms of data validity and up-to-date information. For example, Freebase was shut down in 2016 while OpenCyc's last available version dates back to 2012. We also disregard OpenStreetMaps as other initiatives have thoroughly studied this resource.

\subsubsection*{DBpedia} 
DBpedia is a data source that contains extracted data from Wikipedia info-boxes and that includes abstracts in 11 different languages \cite{paulheim2017knowledge}. According to the latest statistics published (v. 2016-10), DBpedia contained about 6 million entities, including 275K organizations (e.g. 67K companies and 53K educational institutions) \cite{dbpedia2016}. Dbpedia provides mappings to Wikidata, Yago2 and GeoNames as well as numerous other data sources outside the focus of this paper \cite{farber2018linked}. Unfortunately, DBpedia does not provide a unique numerical identifier for a concept, other than its \acrfull{iri}, structured as: \url{http://dbpedia.org/resource/<Title_of_Wikipedia_Article>}. Therefore, we created an identifier for DBpedia items based on their page name (e.g. `Miles\_Laboratories'). 

\subsubsection*{Wikidata} Wikidata is a collaboratively created linked dataset hosted by the Wikimedia Foundation\footnote{\url{https://wikimediafoundation.org/}}. It currently contains about 68 million items (Nov 2019) \cite{wikidatastats} and it is built upon Wikipedia, including different language-specific Wikipedia variants. A concept is identified by a unique number, prefixed with the letter Q, known as a ``QID''. Because of this language-independent identifier, unlike DBpedia, the IRIs of items in Wikidata are consistent among different languages: \url{http://www.wikidata.org/entity/<QID>}. This is an important difference between the two databases since the language-dependent IRI of DBpedia has caused inconsistencies between the property values of items among versions in different languages. 
Wikidata contains mappings to GeoNames and sometimes to \acrshort{umls} or other medical vocabularies like \acrfull{mesh} \cite{lipscomb2000medical}. 

\subsubsection*{Yago} \acrfull{yago} is developed at the Max Planck Institute for Computer Science\footnote{\url{https://www.mpi-inf.mpg.de}}. Just like DBpedia and Wikidata, \acrshort{yago} is also extracted from Wikipedia. Additionally, it uses a categorization system based on WordNet \cite{miller1995wordnet}, and it contains enriched geographic data based on GeoNames. While DBpedia provides links between the different language-specific knowledge graphs, \acrshort{yago} instead aims at multilingual knowledge fusion \cite{paulheim2017knowledge}. This data source is presumed to contain the largest amount of organizations, followed by Wikidata and DBpedia \cite{farber2018linked}. \acrshort{yago}2, the second large iteration of this ontology, provides mappings to DBpedia.

\subsubsection*{Geonames} GeoNames is a geographical database containing over 11 million concepts. It integrates location information such as latitude and longitude coordinate pairs, as well as postal codes, administrative divisions, and countries. GeoNames mostly consists of data provided by official public sources and crowd-sourced modifications to it. GeoNames does not provide a \acrfull{sparql} endpoint to query it, but its ontology is available in OWL while the vocabulary is available as either a \acrshort{rdf} or \acrfull{csv} dump \cite{geonamesstats}. Through its alternative names property, GeoNames provides mappings to Wikidata.

\subsubsection*{UMLS} The \acrfull{umls} is a \acrshort{kb} that unifies many health and biomedical vocabularies and standards. Because of the enormous scope of the \acrshort{umls} Metathesaurus, it reflects the many views of the world present in its source vocabularies \cite{national2009umls}. Therefore, it may be experienced as a rich yet noisy data source. As the \acrshort{umls}  aligns, merges, and relates medical taxonomies and ontologies, it allows researchers and informaticians to navigate the complex biomedical domain and translate concepts across terminologies. Though mostly focused on modeling medical entities (e.g., chemical compounds, diseases, genes), the \acrshort{umls} also contains some information regarding medical organizations. For institutions, under the semantic group `ORGA', the UMLS provides concepts for Healthcare Related Organizations (semantic type T093), Professional Societies (semantic type T094), Self-help, or Relief Organizations (semantic type T095), and a general Organization category (semantic type T092). 
The \acrshort{umls} may provide unique concepts and labels for organizations, but no location data. Yet, due to its importance and heavy adoption in the biomedical domain, mappings to this data source are extremely valuable. 

\subsection*{Data querying and gathering}
\subsubsection*{Query complexity}
The main type of restrictions to cover the three selected medical facility categories are shown in Table \ref{table:queries}.  There is a noticeable difference in the complexity of the SPARQL queries constructed across the three distinct \acrshort{lod} repositories. This makes query portability difficult across the Semantic Web. As for GeoNames and \acrshort{umls},  a filter on the concepts' semantic type (or feature code in GeoNames) is sufficient. 

Beyond the difference across data sources, in some cases retrieving information within a single data source was not straightforward. As studied by \cite{farber2018linked}, one of the challenges when extracting information from DBpedia is the variety in predicates that express the same relation, which inherently drives towards more complex \acrshort{sparql} queries to obtain the needed information. This is particularly evident with geo-location information. For example, to extract the country where an institution is, four different triple patterns exist (Table \ref{table:db_country}).

\subsubsection*{Data validation: facility concepts}\label{sec:validation-facility}
We explicitly blacklist some concepts that do not represent a specific medical facility (see Table \ref{table:queries}, last row). The choice of concepts was driven by heuristics based on preliminary experiments. For example, DBpedia concepts with the name `Hospital' or `University' represent information about the general concept of what a hospital or university is. By using \texttt{rdf:type dbo:Hospital} or \texttt{rdf:type dbo:University} in DBpedia, these entities unintentionally get queried as well because they are specified with these properties. Similarly, UMLS includes items for abstract concepts such as `clinic' that had to be excluded. DBpedia blacklisted items represent 0.02\% of its total retrieved concepts, while \acrshort{umls} items correspond to 0.73\%.

\subsubsection*{Data validation: location of concepts}\label{sec:validation-location} 
Most location information (i.e. country, state or city) in the original data sources are already mapped to GeoNames (123,851 mappings, 32.81\%). However, some location descriptions provided a name but no semantic ID. We attempt to map these to GeoNames as a standardized vocabulary, thus transforming this into semantic information. Textual information was linked by string matching techniques (12,220 mappings, 3.23\%). 2.18\% of location information remains un-linked (i.e., no Geonames id could be retrieved for the given text), and 61.75\% remains unidentified (i.e., NULL values).

Furthermore, in Wikidata no specific information about the state or city is present. Instead, a general `administrative territory' predicate captures this information. Therefore, every territory with a GeoNames ID found in Wikidata was compared against the list of cities (populated places with more than 500 inhabitants) and states (administrative divisions level 1) provided by GeoNames to represent location details in a standardized way.

\subsubsection*{Source coverage and quality}
Figure \ref{fig:source-located} summarizes the coverage per data source: the total number of facility concepts retrieved, and the number of concepts for which we could gather location information at either country, state, city, zip, or geo-coordinates level. Wikidata leads the way by containing the most amount of concepts (61,003), nearly all located at the country level (59,236). However, it suffers from low coverage in the state (6,325) or city (162) levels, arguably due to its joint `administrative territory' property. GeoNames is second place in retrieving concepts (35,788), almost tied with DBpedia (34,415). Finally, \acrshort{yago} takes the third spot (29,742), followed by \acrshort{umls} covering the least amount of concepts (15,954) and no locations.

\subsection*{Alignment and knowledge base construction} 
\label{sec:alignment}
Once the data has been gathered, we merge nodes before adding them to the KB. To merge nodes, the equivalency of two concepts across data sources needs to be established. One way to do so is to take the \texttt{owl:sameAs} mappings most \acrshort{lod} resources provide. Still, the density of the mappings in this sub-domain is poor, as shown in Figure \ref{fig:source-alignment}. Thus, we perform additional alignment as a means to discover new mappings. For simplicity, we enforce a one-to-one mapping among data sources (i.e. one DBpedia concept may correspond to one and only one GeoNames id, one Wikidata id, etc.). 

\noindent The final steps for merging and constructing the KB are:
\begin{enumerate}[\hspace{.5cm}\textbf{Step} 1:]
    \item Add concepts that are \texttt{owl:sameAs} mapped to all other resources in the \acrshort{lod} Cloud. 
    \item Add concepts that are partially \texttt{owl:sameAs} mapped across \acrshort{umls}  Cloud resources.
    \item Add concepts that are mapped by (approximate) string matching. 
    \item Add all unmapped concepts 
\end{enumerate} 

During Step 2 we needed to impose an order over data sources. This decision is important due to the one-to-one mapping requirement. To decide which data source to add first we look at the analysis in Figure \ref{fig:source-located} and select the following order: Wikidata $\rightarrow$ GeoNames $\rightarrow$ DBpedia $\rightarrow$ \acrshort{yago}.  

For Step 3, we first generate candidate entities for aligning a given medical facility by performing an approximate nearest neighbors search. We search over the set of medical facilities' aliases and output the entities associated with the nearest K. Names are encoded using a vector of \acrfull{tfidf} scores of character tri-grams, which appears in 10 or more entity aliases (i.e., document frequency at least 10). In a second processing step, we use the geographical nature of the data and map the concepts heuristically, using three emperically-derived rules, after manual inspection of the data:
\begin{enumerate}
  \item If we know latitude and longitude for both entities, we merge them if their cosine similarity is at least $0.75$ and their geographical distance is at most $1$ km,
  \item If we do not know latitude and longitude for both entities, we merge them if they have the same city and country name and have a cosine similarity of at least $0.9$,
  \item Finally, if we do not know the city name, we merge the entities if they have at least three aliases with a cosine similarity of at least $0.99$ and are within the same country.
\end{enumerate}

\section*{Results} 
\label{sec:evaluation}

\subsection*{Knowledge Base Evaluation: Coverage}

\subsubsection*{Type distribution}  
Table \ref{table:located} shows the distribution of concepts in the KB per semantic type. Healthcare-related organizations (e.g. hospitals) is the category with the most concepts (68,896), followed by Educational Institutions, such as universities (56,667).  

\subsubsection*{Size in nodes} 
We identified 68,896 medical facilities distributed in 202 countries. The contribution and mapping per data source are shown in Figure \ref{fig:source-contribution}. After merging, we observe that we managed to map the majority of concepts into Wikidata, DBpedia and \acrshort{yago} respectively, as can be seen in (Figure  \ref{fig:source-located}). 

\subsection*{Knowledge Base Evaluation: Completeness per node}
\label{sec:completeness}
\subsubsection*{Location}  
Table \ref{table:located} shows the percentage of located concepts in the KB per semantic type. A concept is considered to be located if we have registered a valid value (i.e., not NULL and not (0,0)) for latitude and longitude. Healthcare-related organizations (e.g., hospitals) are the category with the highest percentage of located concepts, which should aid in identifying clinical trial facilities.

\subsubsection*{Language variety}
As per \textbf{Requirement 2}, we have gathered multilingual labels per concept, reaching a total of 115 languages.  Figure \ref{fig:language} compares the total number of concepts and the average number of languages per concept for each data source.  \acrshort{yago} has the highest language coverage, with an average of 2.76 languages per concept. As a result of the alignment, a concept in our  \acrshort{kb} contains 2.09 languages on average. 

\subsection*{Knowledge Base Evaluation: Connectivity} 
\label{sec:connectivity}
We identified 23.43\% mappings across concepts that were not registered in the  \acrshort{lod} Cloud. Figure \ref{fig:source-alignment} illustrates the lack of alignment between datasets (247,201 unidirectional mappings) before the string matching approach, while Figure \ref{fig:our-alignment} shows the result after merging nodes (350,919 unidirectional mappings, resulting in 701,838 bidirectional mappings). As expected, the most benefits are reflected in linking  \acrshort{umls} to  \acrshort{yago} repositories (Figure \ref{fig:our-alignment}, bottom row). Surprisingly, though GeoNames is a standard vocabulary to which DBpedia, Wikidata, and  \acrshort{yago} link when modeling a facility's location, almost no alignment among medical facilities themselves is available. Our approach succeeded in partially mapping GeoNames and other resources, but further efforts are needed. 

The most overlap is observed between DBpedia, Wikidata and  \acrshort{yago} (17,828), followed by the intersection of DBpedia and Wikidata (11,296) (see Figure \ref{fig:source-contribution}).

\subsection*{Knowledge Base Description and Usage}
\label{sec:data} 

\subsubsection*{Modelling}
Figure \ref{fig:schema} shows the schema of our KB. We model four types of semantic resources: medical facilities and three location information levels, i.e., country, state, and city. These semantic resources are associated with data values for name, geo-coordinates and postal code (the latter only being applicable to the clinical trial facility itself). We provide access to the KB as relational data as well as  \acrshort{rdf} triples. 

Relational data is provided as a single  \acrshort{csv} file of 128,506 rows. The  \acrshort{csv} is a flattened representation of the database, where each entry represents a unique concept for a potential clinical facility, associated with its location details (latitude, longitude, city, zip, state, country) and identifiers in the source data. 
 \acrshort{rdf} triples are provided as a turtle file consisting of 1,094,278 statements. The modeling closely follows practices by DBpedia and Wikidata in that it separates vocabulary resources from ontological terms. We use the namespaces \texttt{mto: \url{http://research.mytomorrows.com/ontology/}} and \texttt{mtp: \url{http://research.mytomorrows.com/property/}} to refer to classes and properties in the ontology, while the namespace \texttt{mtr: \url{http://research.mytomorrows.com/resource/}} refers to the specific instances in the vocabulary.

In order to improve the reusability and interoperability of our KB as a linked data resource, our modeling is mapped to SCHEMA.ORG\footnote{\url{http://schema.org/}} and GeoNames. On the one hand, SCHEMA.ORG is selected because the DBpedia, Wikidata, and GeoNames ontologies are consistently mapped to it. On the other hand, GeoNames is selected as the main LOD resource for geographical data. Additionally, resource names are modeled using \acrfull{rdfs}\footnote{\url{http://www.w3.org/2000/01/rdf-schema#}} and \acrfull{skos}\footnote{\url{http://www.w3.org/2004/02/skos/core#}} for preferred and alternative names. Finally, looking back to \textbf{Requirement 1}, we decided to model location semantic resources under the GeoNames conventions, thus assigning a feature code and feature class. 

\subsubsection*{Querying}
The \acrshort{rdf} representation of the KB can be loaded to any triple store. It can be queried as follows:

\begin{lstlisting}[language=sparql]
PREFIX mto: <http://research.mytomorrows.com/ontology/>
PREFIX mtp: <http://research.mytomorrows.com/property/>

SELECT DISTINCT ?site ?site_name ?country_name ?state_name ?city_name
?lat ?long WHERE {
    ?site rdf:type mto:TrialFacility .
    ?site rdfs:label ?site_name .
    
    ?site mtp:country ?country .
    ?country rdfs:label ?country_name .
    
    ?site mtp:state ?state .
    ?state rdfs:label ?state_name .
    
    ?site mtp:city ?city .
    ?city rdfs:label ?city_name .
    
    ?site mtp:latitude ?lat .
    ?site mtp:longitude ?long .
}
\end{lstlisting}

This sample query shows all facility sites in the \acrshort{kb}, including their property country, state, and city. By using \lstinline[columns=fixed]{rdfs:label}, a human-readable version of the resource's name is provided in the result. Furthermore, one can filter the result by country for instance, by adding a specific label to the filter statement in the query as follows: \lstinline[columns=fixed]{FILTER(?country_name="United Kingdom"@en)}.

We can also exploit the geographical properties of our \acrshort{kb} using GEOSPARQL. The recommended triple stores are GraphDB or Stardog as they support GEOSPARQL. The following query retrieves information about trial facilities and their distance to a given point.

\begin{lstlisting}[language=sparql]
PREFIX rdf: <http://www.w3.org/1999/02/22-rdf-syntax-ns#>
PREFIX mto: <http://research.mytomorrows.com/ontology/>
PREFIX geo: <http://www.opengis.net/ont/geosparql#>
PREFIX geof: <http://www.opengis.net/def/function/geosparql/>
PREFIX uom: <http://www.opengis.net/def/uom/OGC/1.0/>
PREFIX mtp: <http://research.mytomorrows.com/property/>
PREFIX rdfs: <http://www.w3.org/2000/01/rdf-schema#>

SELECT DISTINCT ?site ?distance ?country ?state ?city ?coor WHERE {
  ?site rdf:type mto:TrialFacility;
    geo:asWKT ?coor.
    
  BIND("POINT(52.381683 4.807050)"^^geo:wktLiteral AS ?mycoor)
  BIND(geof:distance(?coor, ?mycoor, uom:metre) AS ?distance)
  
  OPTIONAL {
    ?site mtp:country ?co.
    ?co rdfs:label ?country.
  }
  OPTIONAL {
    ?site mtp:state ?st.
    ?st rdfs:label ?state.
  }
  OPTIONAL {
    ?site mtp:city ?ci.
    ?ci rdfs:label ?country.
  }
}
ORDER BY (?distance)
\end{lstlisting}

\section*{Discussion}
\label{sec:alt-use-case}
Our \acrlong{kb} constitutes two main contributions in terms of data and the method via which it is obtained, i.e. medical facility coverage and recency, and facilities identified as clinical trial sites. 

\subsection*{Knowledge Base Coverage and Recency}
\label{sec:coverage} 
Through the merging of various frequently updated individual data sources such as Wikidata, our knowledge base offers recent data and more extensive coverage of facilities than that of any of the individual sources used. Considering that the \acrfull{who} has last reported the number of medical facilities per 100,000 inhabitants in every country in 2013/2014 \cite{whoreport}, recency and coverage may be quite valuable to those who rely on this \acrshort{who} data.

Our knowledge base combined, for instance, with disease incidence data per country \cite{ballas2006radiation} or with hospital capacity (a property present in some LOD resources), may also offer a starting point for policymakers to make better-informed decisions regarding the local organization of healthcare.

As illustrated in Figures \ref{fig:WHO} and \ref{fig:KB} showing a geographical representation of hospitals compared to the \acrshort{who}, our knowledge base is higher in coverage. However, what comes with this gain is a lower level of validation of the data, which needs to be taken into consideration when using the knowledge base. For instance, our exact definition of what a hospital is may differ significantly from that of the WHO. Indeed, some countries in the \acrshort{who} dataset only consider public hospitals, and others also include private medical facilities, while our \acrshort{kb} includes both private and public hospitals to a greater extent. As shown in the figures linked above, hospital density information and other knowledge covered by our knowledge base should be critically examined upon usage.

\subsection*{Sites linked to clinical trials}
\label{sec:triallinks} 
By connecting facilities to clinical trials listed in \acrshort{ctgov} using various approaches, we believe we may be facilitating several critical use cases, some of which we are discussing below, backed up by data from our knowledge base whenever we are able to do so.

First of all, a search engine's ability to connect trials to facility entities may be a tremendous help for patients in need. Better linkage may enable the retrieval/filtering of trials that are within a specified range (such as within 50km from someone's home) with higher recall (as opposed to having relevant trials not being returned because their location string is not linked to a known entity with latitude/longitude/city data). Also, linked data may offer fewer false positives, resulting from loose matching of strings employed in various current trial search engines. 

As a proof-of-principle, we performed entity linking of our \acrshort{kb} to a snapshot of clinical trials collected from \acrshort{who} on January 12\textsuperscript{th} 2021. The download contains 309,658 trials where at least one trial site is mentioned, with an average of 7 trial sites per clinical trial and 871,570 distinct combinations of clinical trial site name and location (city, zip, state, country). We link the clinical trial sites to our KB using an approximate nearest neighbor algorithm, similar to the (approximate) string matching approach described for aligning concepts in our \acrshort{kb}. However, this time, we filter matches on cosine similarity (greater than 0,7) and location. We create 206,218 one-to-one matches between distinct clinical trial sites and medical facilities in our KB, corresponding to 46\% (1,027,955) of the clinical trial sites mentioned in \acrshort{ctgov}.

Table \ref{table:trialsites}  gives an overview of the top 5 medical facilities linked to the most clinical trials and the top 5 clinical trial site mentions we failed to match. An interesting occurrence is that of a pharmaceutical company (Boehringer Ingelheim) matched to 4,062 trials as a clinical trial site. Upon closer inspection of the dataset, this is an example of a particular phenomenon that can be found in clinical trial documents:  
For several trial documents, clinical trial site names are vague, either because they are generic (e.g. ``Hospital'') or because they are intentionally hidden (e.g. ``Novartis Facility \#1''). String matching based approaches will not identify these locations correctly, excluding these sites from query results. To tackle this, we created specific patterns to identify vague clinical trial site names, as shown in Table \ref{table:vague_patterns}. These patterns identify 75,664 unique clinical trial site names (16.56\% of the total clinical trial site names, and 21.81\% of the unidentified clinical trial site names). Since the names described by these patterns do not represent specific medical facilities, these clinical trial site names must follow alternative linking approaches that go beyond the KB medical facilities' names quality. An extension to our work would be to look into deanonymizing these locations for our use cases using clinical trial sites' location properties and those of medical facilities in our KB.

As an illustration of the use of such a semantic clinical trial site linking, we performed a location-based search in four of the largest clinical trial search engines: \acrshort{ctgov}, \acrshort{eudra}, \acrshort{ictrp}, and a clinical trial search engine powered by our generated semantic database. As a query, we aim to find trials running in the ``University of Michigan''. Screenshots of the search engines are provided in Figure \ref{fig:search-michigan}. \acrshort{ctgov} retrieves the most related studies (3,211) to these keywords. However, the results are not precise as they relate to each token independently and retrieve studies performed in any `University'. On the opposite side of the spectrum, \acrshort{eudra} retrieves the lowest amount of documents (16), including trials that are only running in the United Kingdom but are retrieved due to similar reasons as \acrshort{ctgov}. \acrshort{ictrp} lies in the middle range in retrieving relevant trials (1,176) together with our search engine (1,120). Upon close examination, we see that the \acrshort{ictrp} results include studies sponsored by the University of Michigan but taking place in different locations. We do not have the ground-truth to validate the accuracy (precision/recall) of the retrieval, but still, given the `findability' aspect of our use case, we may see that bringing semantics into this domain would provide clarity regarding what each entity represents and the role each entity has in the clinical trial.

A knowledge base such as ours may also be helpful while choosing suitable clinical trial locations. Looking at accurate historical records of clinical trials at locations, as well as locating expertise via such data, may be a valuable supporting step for this choice. Furthermore, it has been established that a large percentage of trials fail due to the inability to recruit patients, rather than clinical reasons such as toxicity or the lack of efficacy of the medicine \cite{stensland2020estimating, bandari2020clinical}. A major reason for patients to be rejected from clinical trial participation is that they live too far away from a trial site \cite{ellis2020patient}. Thus, a high-coverage, recent knowledge base tied to past or current clinical trials -- perhaps combined with other relevant data sources -- may ultimately lead to an efficiency increase in the drug development process.

\section*{Conclusions}
\label{sec:conclusions} 
The \acrshort{lod} initiative has the vision of making data accessible and interlinked (``With linked data, when you have some of it, you can find other, related, data.''\cite{timlod}). Yet, in the domain of location of medical sites the problem of poor connectivity limits the usability of any of these resources. This could mean that we are not serving society well, since significant effort is needed to identify medical facilities thus interfering with a basic human right: access to health care. 

While tackling the challenge of creating a unified \acrshort{kb} for medical facilities, we learned several lessons:
\begin{enumerate}
    \item Extensive exploration is needed to query each individual data source due to the lack of query portability across repositories.
    \item The set of mappings across linked data repositories is not complete; and the coverage of location data on each repository differs greatly. These two factors present the perfect use case for ontology alignment benefits to be maximized. On the one hand, distinct nodes increase the overall coverage of the knowledge graph. On the other hand, overlapping nodes enrich each other by containing different information per source (e.g., labels in different languages and different location information).  
    \item When data sources do not separate ontological concepts from vocabulary resources, some general entities like `University' may be wrongly fetched. A `blacklisting' step benefits the data quality of the KB. The items to blacklist are gathered during the exploration phase for every data source.
    \item Counter-intuitively, aligning resources may hurt some \acrshort{kb} evaluation metrics. For example, we observe a compromise between increasing KB coverage (adding more nodes) and decreasing completeness (the lack of information available per node). 
\end{enumerate}

Our analysis may shed some light on future research efforts. For example, our connectivity analysis illustrates that there is still work to be done on linking GeoNames concepts to other \acrshort{lod} sources, specifically outside the country/city semantic types. On a similar note, our completeness analysis brings to attention that Wikidata has a surprisingly low language coverage in this domain, despite its efforts for being a highly multilingual \acrshort{lod} source. 

One of our \acrshort{kb}'s largest limitations is the validity of the data, as it directly depends on the quality of the original data sources.  We performed in-house expert validation due to the access to health care professionals we have in our lab, but an approach as in-depth as \cite{ballas2006radiation} is extremely costly. In the future, we could leverage the acquired knowledge to design more general and less expensive techniques. Furthermore, the current implementation of the medical facilities \acrshort{kb} is heavily focused towards gathering multilingual labels and location properties. In the future, we plan on accounting for relations between organizations for example a hospital as part of a university (e.g. Academic Medical Center and University of Amsterdam). \\

\noindent To conclude, in this paper we have described the generation of a \acrshort{kb} following the identification and alignment of several data sources containing medical facility information. We believe that both the resulting \acrshort{kb} and the lessons learned while creating it might be valuable for researchers and practitioners needing this, or similar biomedical resources. 

\printglossary

\printglossary[type=\acronymtype, title=List of abbreviations]


\begin{backmatter}
\section*{Declarations}

\section*{Ethics approval and consent to participate}
Not applicable

\section*{Consent for publication}
Not applicable
  
\section*{Availability of data and materials}
The datasets generated and/or analysed during the current study are available in the myTomorrows repository, \url{https://research.mytomorrows.com/datasets}.

\section*{Competing interests}
  The authors declare that they have no competing interests.

\section*{Funding}
Not applicable

\section*{Author's contributions}
Data gathering: GB, LPT, SBS. Alignment: SBS, RJS. Evaluation design: ZS. Evaluation: EM, SBS.

\section*{Acknowledgements}
Not applicable
    

\bibliographystyle{bmc-mathphys} 
\bibliography{bmc_article}      




\section*{Figures}

\begin{figure}[!ht]
\centering
\includegraphics[width=.95\textwidth]{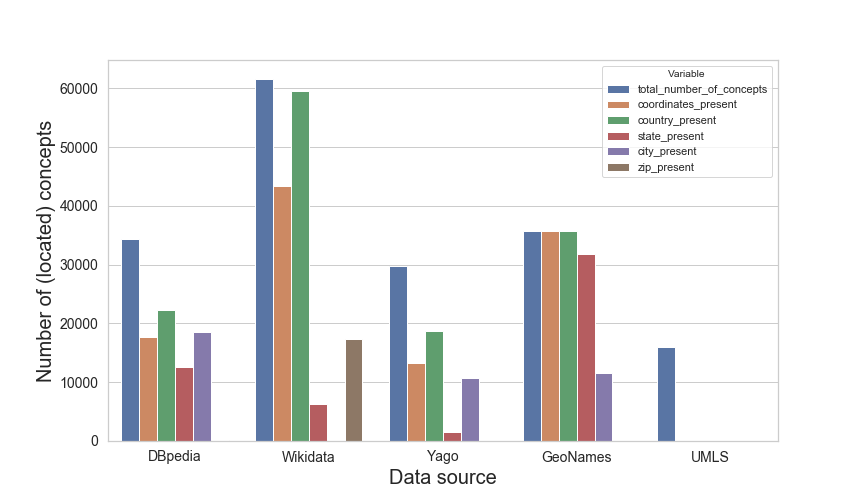}
    \caption{\label{fig:source-located} Number of concepts and locations per source. GeoNames is the most complete data source with geo-coordinates for all concepts and country location, while UMLS contains no location data.}
\end{figure}

\begin{figure}[!ht]
\centering
\includegraphics[width=.9\textwidth]{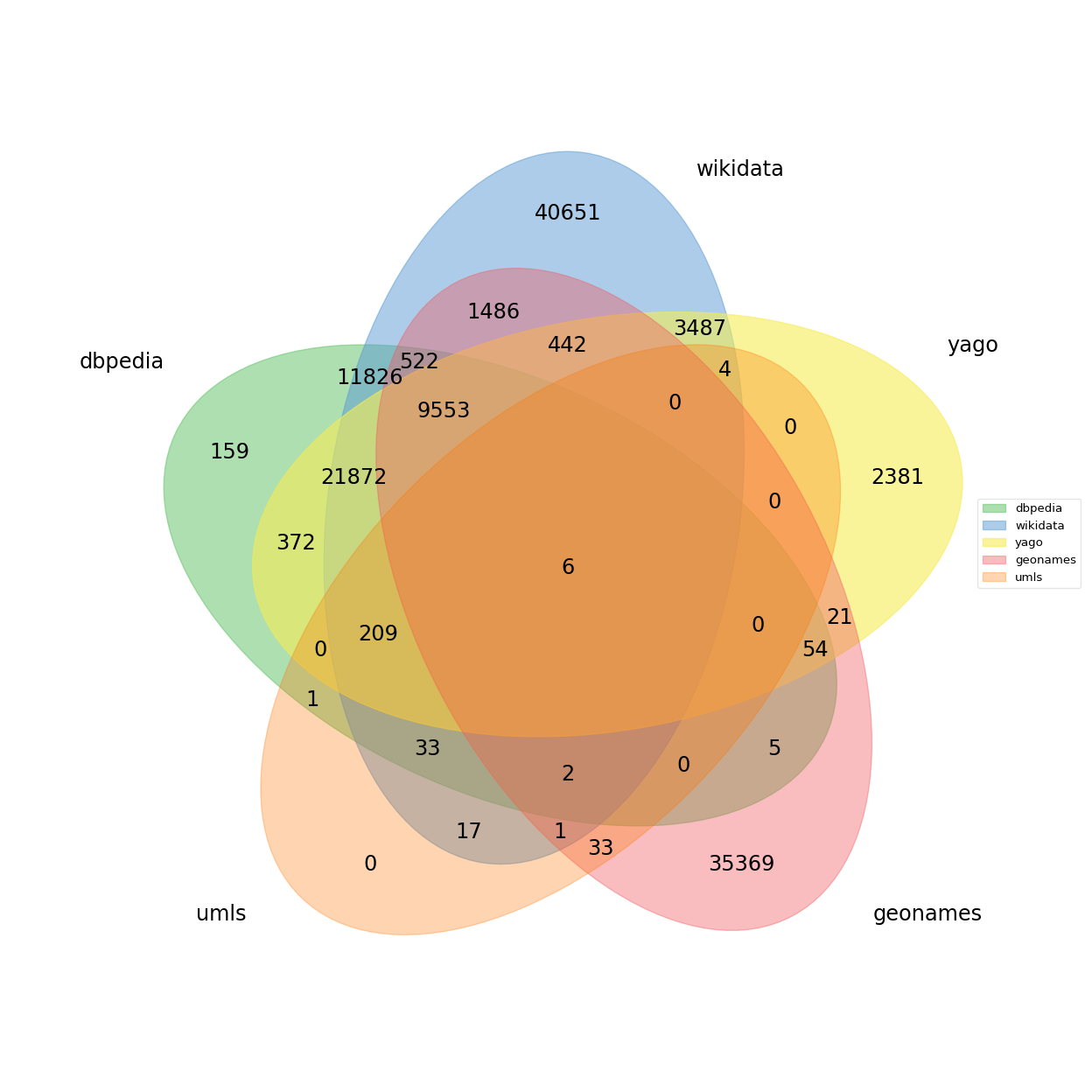}
    \caption{\label{fig:source-contribution} Final concept contribution per data source to merged Knowledge Base.}
\end{figure}
\begin{figure}[!ht]
\centering
\includegraphics[width=.95\textwidth]{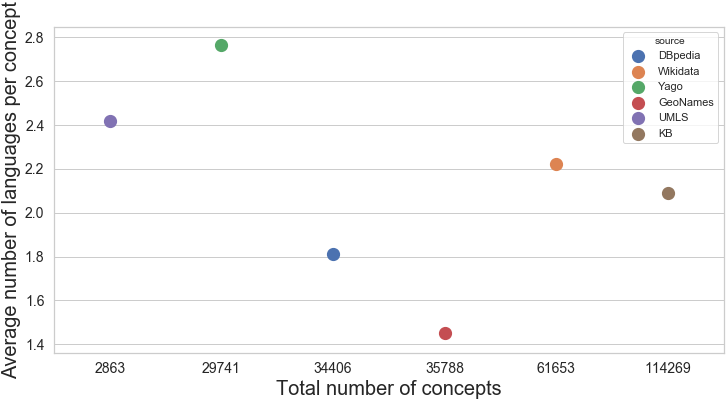}
    \caption{\label{fig:language} Average number of languages in alternative names per concept in the Knowledge Base.}
\end{figure}

\begin{figure}[!ht]
\centering
\begin{minipage}{.48\textwidth}
  \centering
  \includegraphics[width=\linewidth]{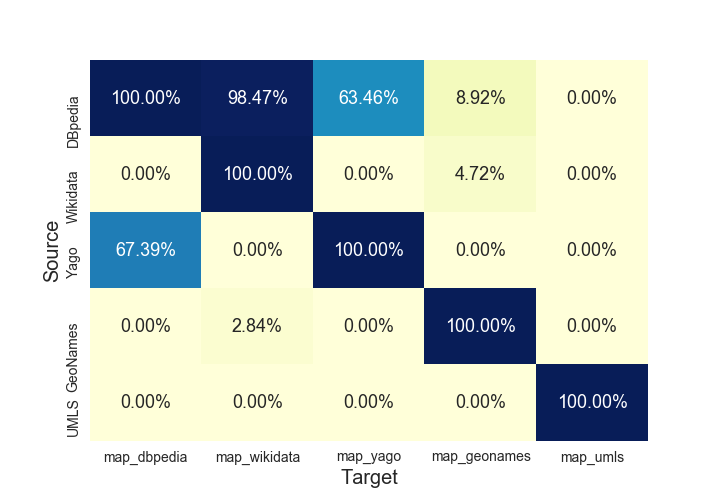}
  \caption{Alignment across source databases}
    \label{fig:source-alignment}
\end{minipage}%
\begin{minipage}{.48\textwidth}
  \centering
    \includegraphics[width=\linewidth]{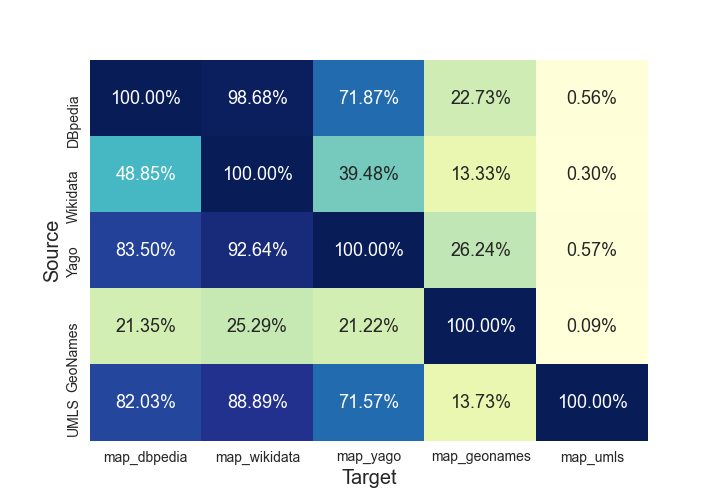}
  \caption{Alignment achieved by our approach}
    \label{fig:our-alignment}
\end{minipage}
\end{figure}

\begin{figure}[!ht]
\centering
\includegraphics[width=.8\textwidth]{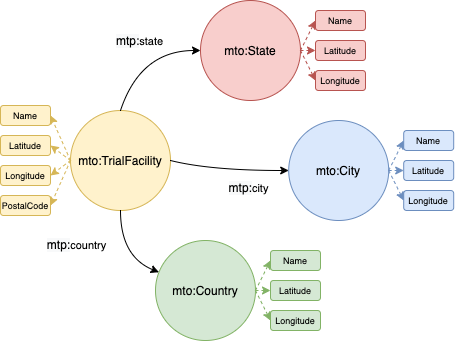}
    \caption{TBox for resulting KB.}
    \label{fig:schema}
\end{figure}

\begin{figure}[!ht]
\centering
\begin{minipage}{.48\textwidth}
  \centering
  \includegraphics[width=.95\linewidth]{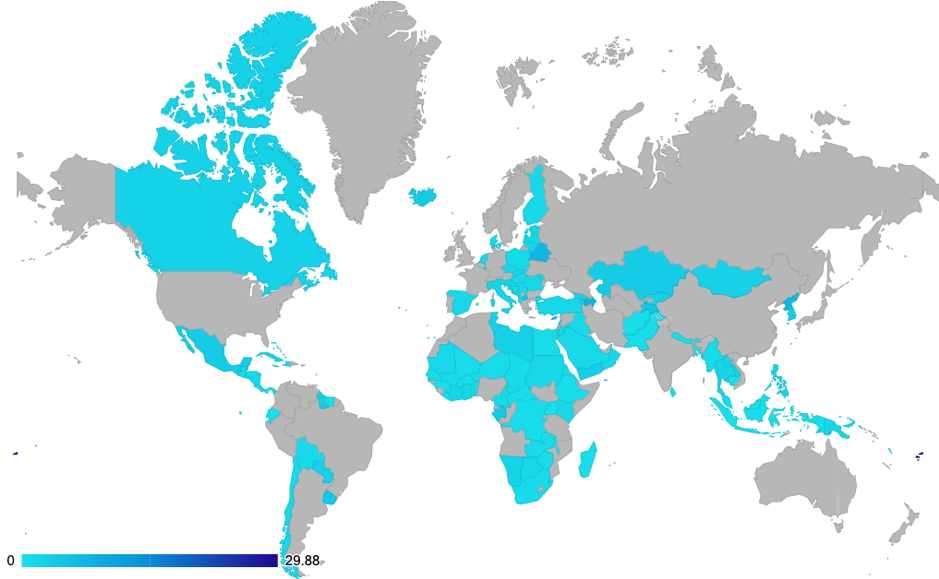}
  \caption{Hospital density per 100,000 \\ inhabitants. Source: WHO data. \\ Countries with no data are \\ marked in grey.}
    \label{fig:WHO}
\end{minipage}%
\begin{minipage}{.48\textwidth}
  \centering
    \includegraphics[width=.95\linewidth]{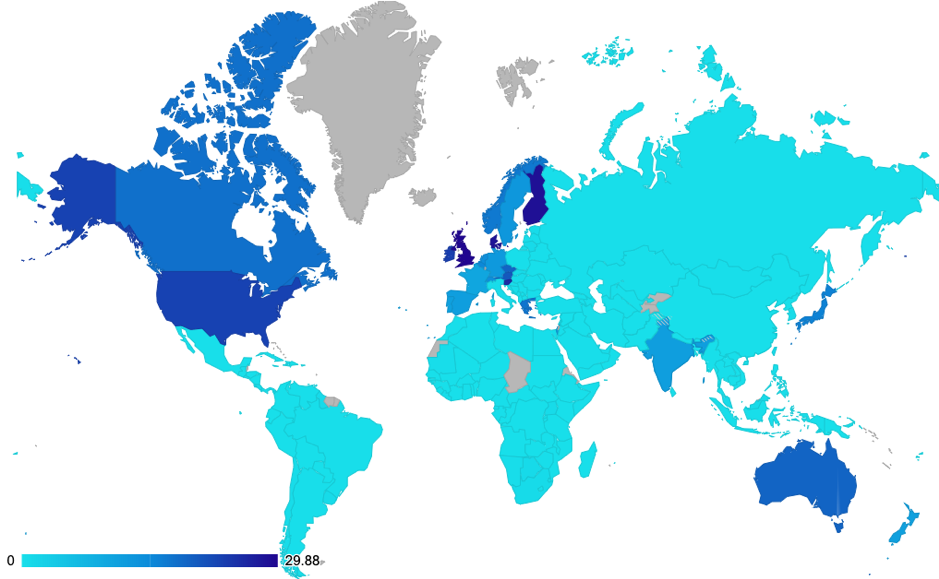}
  \caption{Hospital density per 100,000 \\ inhabitants. Source: Our KB. \\ Countries with no data, or a population of less than 500,000 are marked in grey.}
    \label{fig:KB}
\end{minipage}
\end{figure}

\begin{figure}[!ht]
\centering
\begin{minipage}{.7\textwidth}
  \centering
  \includegraphics[width=\linewidth]{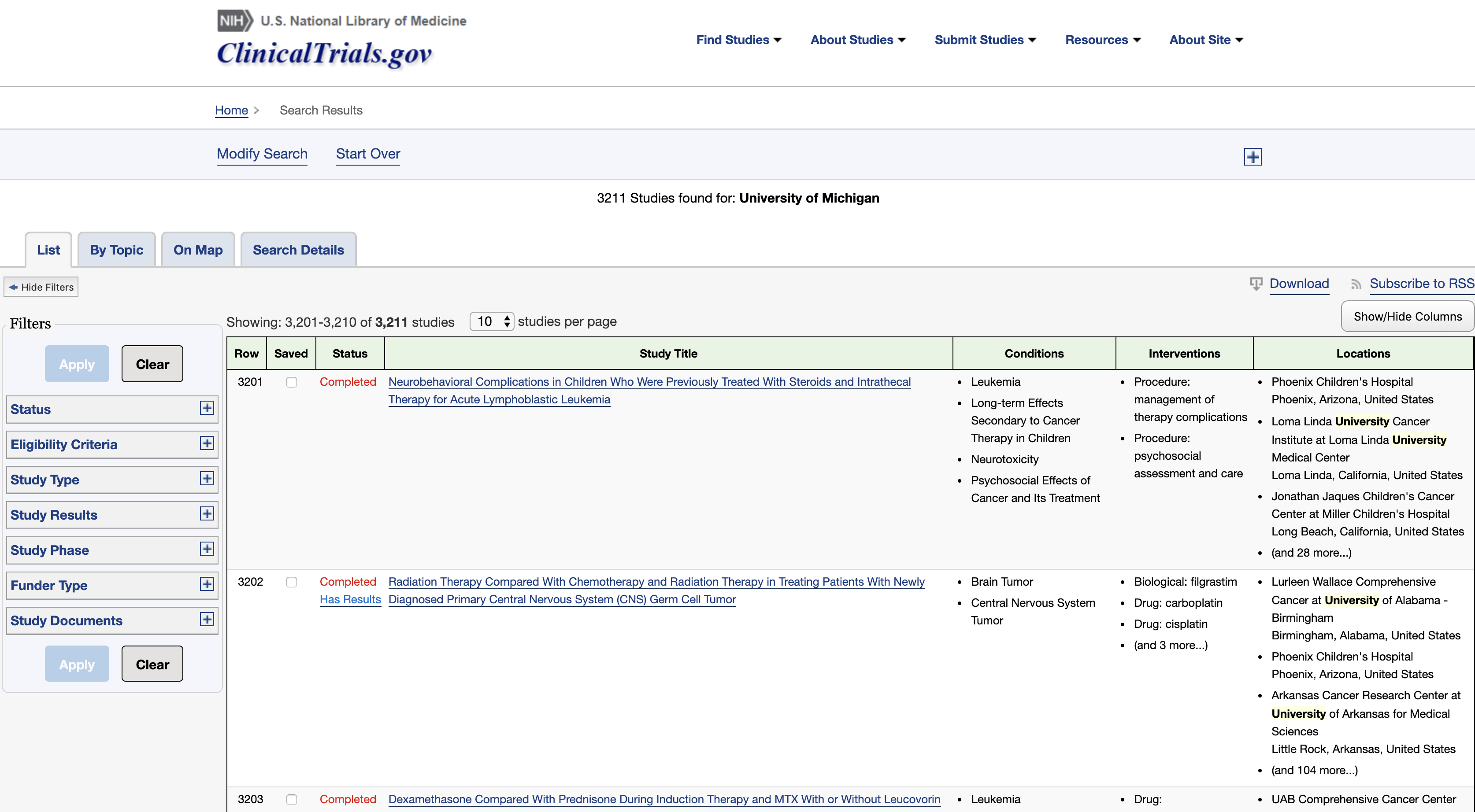}
\end{minipage}%
\\
\begin{minipage}{.7\textwidth}
  \centering
  \includegraphics[width=\linewidth]{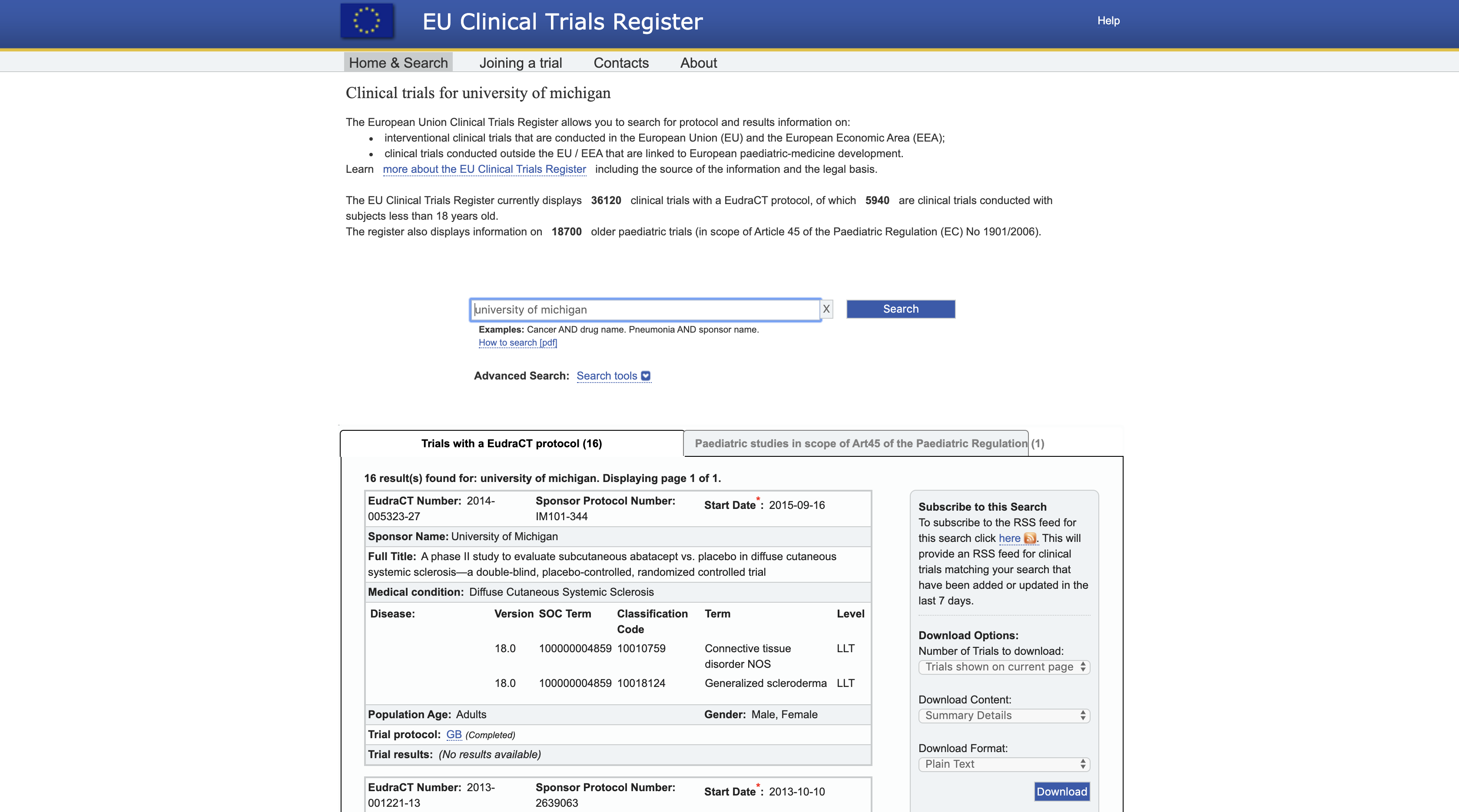}
\end{minipage}
\\
\begin{minipage}{.7\textwidth}
  \centering
  \includegraphics[width=\linewidth]{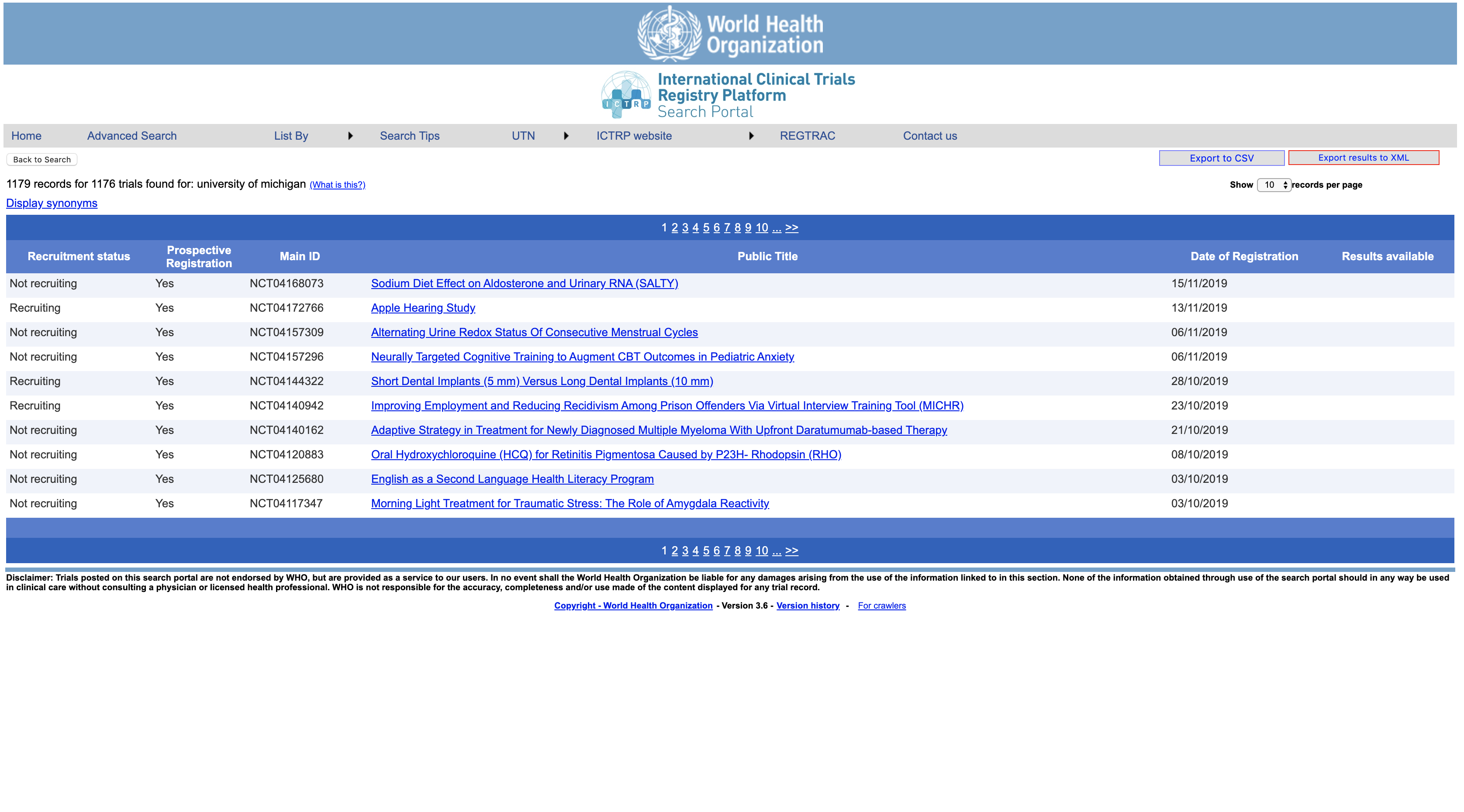}
\end{minipage}%
\\
\begin{minipage}{.7\textwidth}
  \centering
  \includegraphics[width=\linewidth]{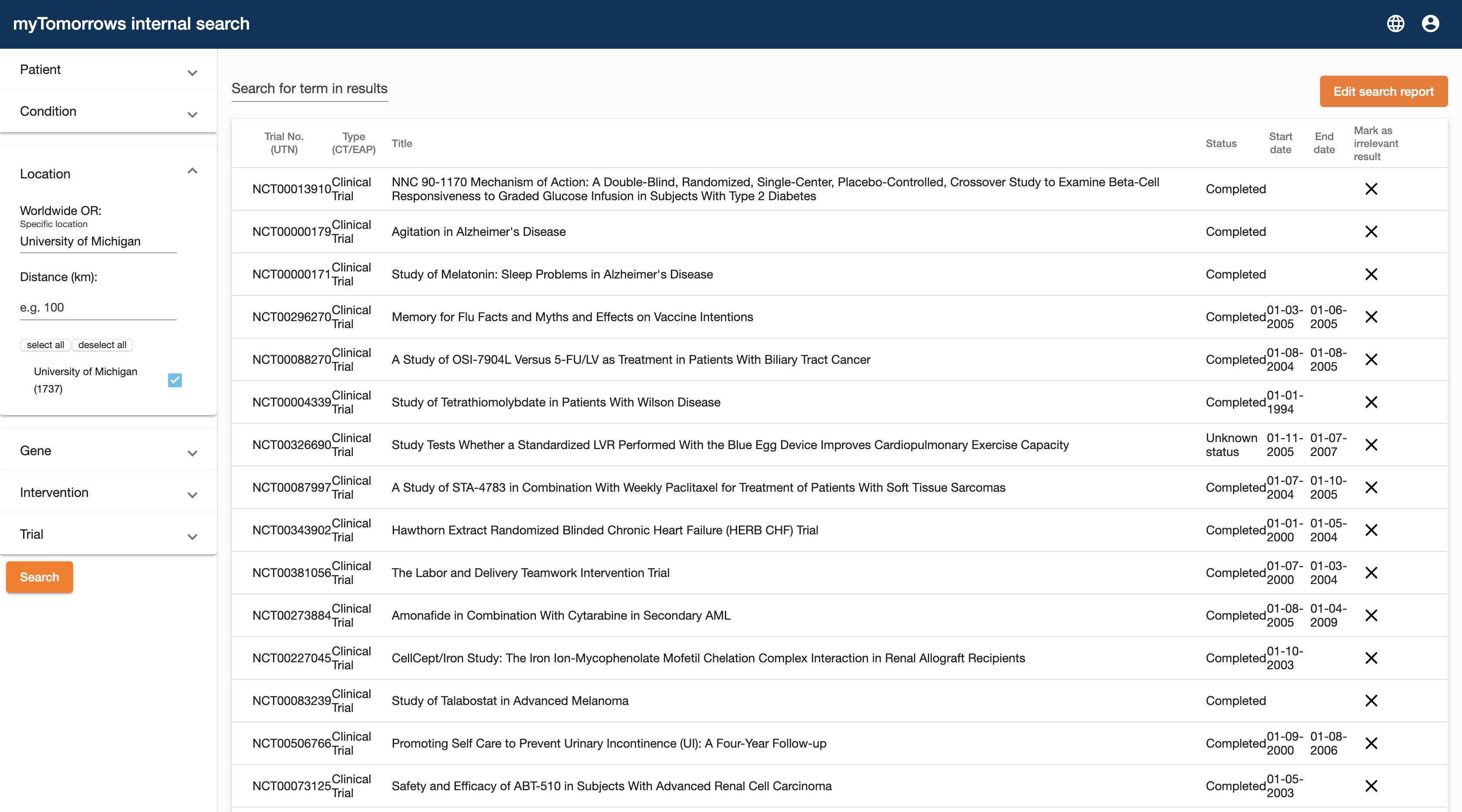}
\end{minipage}
\caption{Comparison of search results and interfaces when attempting to retrieve clinical trials or early access programs happening in the University of Michigan. From top to bottom: CTGOV, EUDRA, ICTRP, our search engine. Screenshots were produced in October 2019.}
\label{fig:search-michigan}
\end{figure}



\newpage
\section*{Tables}

\begin{table}[!ht]
\centering
\caption{Triple patterns to retrieve country information in DBpedia}
\begin{tabular}{p{2.5cm} p{3.5cm} p{2.5cm}}\hline
?item & dbo:locationCountry & ?country \\\hline
?item & dbo:country & ?country \\\hline
?item & dbp:country & ?country \\\hline
?item & dbo:location & ?location . \\
?location & dbo:country & ?country \\\hline
\end{tabular}
\label{table:db_country}
\end{table}  

\begin{sidewaystable}[!ht]
\centering
\caption{Types of entities retrieved across sources}
\begin{tabular}{|c||p{2cm}|p{3.8cm}||p{1.6cm}|p{1.9cm}||p{1.2cm}|p{4cm}||p{1.4cm}||p{2.7cm}|}\hline
 & \multicolumn{2}{|c|}{DBpedia} & \multicolumn{2}{|c|}{Wikidata} & \multicolumn{2}{|c|}{Yago} & GeoNames & \multicolumn{1}{|c|}{UMLS} \\\hline
 & ?predicate & ?type & ?predicate & ?type & ?predicate & ?type & Feature code & Semantic Type  \\\hline

\hline \hline
 
{\rotatebox[origin=c]{90}{\specialcell{Educational \\Institutions}}} & \url{rdf:type} & \specialcell{\url{dbo:University},\\ \url{yago:University108286569},\\ \url{yago:WikicatUniversities}\\\url{AndColleges},\\ \url{schema:CollegeOrUniversity},\\ \url{umbel-rc:University}} & \specialcell{\url{wdt:P31 }\\\url{/ wdt:P279*}} & \url{wd:Q38723} & \url{rdf:type} & \url{yago:wordnet_university_108286569} & \url{S.UNIV}, \url{S.UNIP}, \url{S.ITTR}, \url{S.STNB} &  \\\hline

\hline \hline

\multirow{3}{*}{\rotatebox[origin=c]{90}{\specialcell{Hospitals \\and Medical Centers}}} & \url{rdf:type} & \specialcell{\url{dbo:Hospital},\\ \url{yago:Hospital103540595},\\ \url{schema:Hospital},\\ \url{yago:MedicalBuilding}\\\url{103739518},\\ \url{yago:WikicatUniversity}\\\url{Hospitals}} & \specialcell{\url{wdt:P31 }\\\url{/ wdt:P279*}} & \url{wd:Q16917}, \url{wd:Q1774898}, \url{wd:Q31855} & \url{rdf:type} & \specialcell{\url{yago:wordnet_medical}\\\url{_building_103739518},\\ \url{ yago:wordnet_hospital}\\\url{_103540595},\\ \url{yago:wordnet_clinic}\\\url{_103043274}} & \url{S.HSP}, \url{S.HSPC}, \url{S.HSPD}, \url{S.HSPL}, \url{S.CTRM} & \url{T092}, \url{T093}, \url{T094}, \url{T095} \\\cline{2-9}
  & \url{dct:subject } & \specialcell{\url{dbc:Hospital_networks},\\ \url{dbc:Medical_and_health}\\\url{_organizations},\\ \url{dbc:Clinics}} & \url{?rel} & \specialcell{\url{[wdt:P31}\\\url{wd:Q12136]}} & & &  &  \\\cline{2-9}
  & \url{dbo:industry} & \url{dbr:Health_care} & & & & & & \\\hline
 
\hline \hline
 
\multirow{3}{*}{\rotatebox[origin=c]{90}{\specialcell{Pharmaceutical \\ companies}}} & \url{rdf:type} & \url{yago:DrugCompany108002578} & \specialcell{\url{wdt:P31 }\\\url{/ wdt:P279*}} & \url{wd:Q4830453} & \url{rdf:type} & \specialcell{\url{yago:wikicat_Biotechnology}\\\url{_companies},\\ \url{yago:wikicat_Pharmaceutical}\\\url{_companies},\\ \url{yago:wordnet_drug_company}\\\url{_108002578}} &  &  \\\cline{2-9}
 &  &  & \url{wdt:P452} & \url{wd:Q507443}, \url{wd:Q7108} & & & &  \\\cline{2-9}
 &  &   & \url{?rel} &  \url{wd:Q507443}, \url{wd:Q15067276} &  & & &  \\\hline
 
\hline \hline
 
\multirow{1}{*}{\rotatebox[origin=c]{90}{\specialcell{Blacklisted \\items}}} & \url{IRI} & \specialcell{\url{Hospital},\\ \url{Rehabilitation_hospital},\\ \url{University},\\ \url{School},\\ \url{Community_college},\\ \url{Sixth_form_college},\\ \url{Professional_school},\\ \url{Reform_school}} & & & & &  & \specialcell{C3828423, C0557812, \\C0027689, C0681333, \\C1318228, C0871230, \\C0337962, C0036375, \\C0557808, C1562642, \\C0598858, C0587907, \\C3833703, C3838700, \\C0020021, C0442592, \\C3839733, C3810847, \\C3840771, C0237680, \\C3834491} \\\hline
\end{tabular}
\label{table:queries}
\end{sidewaystable}

\begin{table}[!ht]
\centering
\caption{Percentage of entities with latitude and longitude.}
\begin{tabular}{|p{4.3cm} | p{2.2cm} | p{2.2cm} | p{2.2cm}|}\hline
\textbf{Semantic type} & \textbf{Total concepts} & \textbf{Concepts with coordinates} & \textbf{Percentage with coordinates } \\\hline\hline
Educational Institution & 47,716 & 25,290		& 53,01 \% \\\hline
Healthcare Related Organization & 64,648 & 58,374 & 90,30\% \\\hline
Pharmaceutical Company & 1,643 & 38 & 2.32\%  \\\hline
\end{tabular}
\label{table:located}
\end{table} 

\begin{table}[!ht]
\centering
\caption{Top 5 facilities and top 5 unmatched sites ranked by amount of trials.}
\begin{tabular}{|p{5cm} | p{1cm} | p{5cm} | p{1cm}|}\hline
\textbf{Facility} & \textbf{Trials} & \textbf{Site mention} & \textbf{Trials } \\\hline\hline
Case Comprehensive Cancer Center & 7,466 & National Institutes of Health Clinical Center, 9000 Rockville Pike & 2,284 \\\hline
MD Anderson Cancer Center & 5,358 & Seoul National University Hospital & 562 \\\hline
Memorial Sloan Kettering Cancer Center & 5,036 & Henry Ford Health System & 465 \\\hline
General Hospital & 4,320 & Asan Medical Center & 432 \\\hline
Boehringer Ingelheim & 4,062 & Samsung Medical Center & 409 \\\hline
\end{tabular}
\label{table:trialsites}
\end{table} 

\begin{table}[!ht]
\centering
\caption{Coverage per pattern to discover vague clinical trial sites. The symbol: \textbf{\%} represents a wildcard}
\begin{tabular}{|p{3cm}|p{1.3cm}|p{1.3cm}|p{5.3cm}|}\hline
\textbf{Pattern} & \textbf{Total number of clinical trial sites} & \textbf{Number of distinct name patterns} & \textbf{Examples} \\\hline\hline
`\%Investigational Site\%' & 80,678 & 45,923 & `Pfizer Investigational Site', `Novartis Investigational Site' \\\hline
`\%Facility \#\%' & 71 & 21 & `Facility \# 1', `Facility \#1' \\\hline
`\%Investigative Site\%' & 10,312 & 3,175 & `For additional information regarding investigative sites for this trial, contact the Clinical Trials Support Center at 1-877-CTLILLY (1-877-285-4559) or speak with your personal physician.' \\\hline
`\%Hospital \#\%' & 1,691 & 1,103 & `Moscow City Hospital \# 29', `St- Petersburg GUZ City Hospital \#15', `City Clinical Hospital \#8', `NEI HPE Moscow State University of Medicine and Dentistry of FAHSD City Clinical Hospital \#70', `City Hospital \#31',  \\\hline 
`\%Research Site\%' & 5,400 & 1,676 & `Clinical Research Site', `Research Site', `Research Sites', `Shire Clinical Research Site', `Research Site 1', `Furiex Research Site', `CAPRISA, Vulindlela Clinical Research Site' \\\hline
`Site\%' & 33,370 & 23,577 & `Site 128', `Site \# 3 OLVG Amsterdam' \\\hline
`Local Ins\%' & 933 & 102 & `Local Institution', `Local Instiution', `Local Instiuttion', `Local Institution - Houston', `Local Institute'  \\\hline
`Local Inv\%' & 3 & 1 & `Local Investigator'\\\hline
`Local Hosp\%' & 3 & 2 & `Local Hospitals', `Local hospital at the station Smolensk OAO RZD' \\\hline
`Hospital Local\%' & 3 & 3 & `Hospital Local Palos Blancos', `Hospital local de Yopal E.S.E', `Hospital Local' \\\hline
`\%local\%office\%' & 14 & 6 & `Local Medical Information Office', `Pfizer Local Country Office', `US Local Med Info Office', `Local US Medical Information Office', `Local Medical Information Office', `Pfizer Japan Local Country Office' \\\hline
`\%local\%information\%' & 14 & 11 & `Local Medical Information Office', `Pfizer Local Country Office', `US Local Med Info Office' \\\hline
`\%local\%authority of\%' & 11 & 11 & `Local Health Authority of Reggio Emilia-IRCCS', `Uzsoki Hospital of the Local Authority of Budapest, Oncoradiology', `Local Heath Authority of Padova', `Local Heath Authority of Verona', `Local Health Authority of Modena', `Local Health Authority of Pieve di Soligo', `Local Health Authority of Mirano', `Local Heath Authority of Thiene' \\\hline
`Azienda\%Sanitaria Locale\%' & 78 & 56 & `Azienda Unità Sanitaria Locale di Reggio Emilia', `Azienda Sanitaria Locale Brindisi', `Azienda Unità Sanitaria Locale-IRCCS - Arcispedale Santa Maria Nuova, Ematologia' \\\hline
`Local primary school\%' & 5 & 1 & `Local primary schools'\\\hline
\end{tabular}
\label{table:vague_patterns}
\end{table} 




\end{backmatter}
\end{document}